\documentclass[a4paper,fleqn]{cas-dc}

\usepackage[numbers]{natbib}
\usepackage{hyperref}
\usepackage{subfig}
\usepackage{placeins}

\def\tsc#1{\csdef{#1}{\textsc{\lowercase{#1}}\xspace}}
\tsc{WGM}
\tsc{QE}
\tsc{EP}
\tsc{PMS}
\tsc{BEC}
\tsc{DE}

\graphicspath{{./figs/}}

\begin{document}
\let\WriteBookmarks\relax
\def\floatpagepagefraction{1}
\def\textpagefraction{.001}
\shorttitle{PIV-FlowDiffuser:Transfer-learning-based denoising diffusion models for PIV}
\shortauthors{Zhu et~al.}

\title [mode = title]{PIV-FlowDiffuser: Transfer-learning-based denoising diffusion models for particle image velocimetry}                      



\author[1]{Qianyu Zhu}[style=chinese]
\credit{Conceptualization of this study, Methodology, Model Training, Writing - Original draft preparation,}


\author[1]{Junjie Wang}[style=chinese]
\credit{Conceptualization of this study, Data curation, Writing - Original draft preparation}

\author[1]{Jeremiah Hu}[style=chinese]
\credit{Model Training, Writing - Original draft,}

\author[1]{Jia Ai}[style=chinese]
\credit{Methodology, Writing - Original draft,}

\author[1]{Yong Lee}[style=chinese]
\cormark[1]
\ead{yonglee@whut.edu.cn}
\credit{Conceptualization of this study, Methodology, Writing}

\affiliation[1]{organization={Hubei Provincial Engineering Research Center of Robotics \& Intelligent Manufacturing\\
School of Mechanical and Electronic Engineering, Wuhan University of Technology (WHUT)},
addressline={122 Luoshi Road}, 
postcode={430070}, 
postcodesep={}, 
city={Wuhan},
country={China}}

\begin{abstract}
Deep learning algorithms have significantly reduced the computational time and improved the spatial resolution of particle image velocimetry~(PIV). However, the models trained on synthetic datasets might have a degraded performance on practical particle images due to domain gaps. As a result,  special residual patterns are often observed for the vector fields of deep learning-based estimators. To reduce the special noise step-by-step, we employ a denoising diffusion model~(FlowDiffuser) for PIV analysis. And the data-hungry iterative denoising diffusion model is trained via a transfer learning strategy, resulting in our PIV-FlowDiffuser method. Specifically, (1) pre-training a FlowDiffuser model with multiple optical flow datasets of the computer vision community, such as Sintel, KITTI, etc;
(2) fine-tuning the pre-trained model on synthetic PIV datasets. 
Note that the PIV images are upsampled by a factor of two to resolve the small-scale turbulent flow structures. The visualized results indicate that our PIV-FlowDiffuser effectively suppresses the noise patterns. Therefore, the denoising diffusion model reduces the average end-point error~($AEE$) by $59.4\%$ over RAFT256-PIV baseline on the classic Cai's dataset. Besides, PIV-FlowDiffuser exhibits enhanced generalization performance on unseen particle images due to transfer learning. Overall, this study highlights the transfer-learning-based denoising diffusion models for PIV.  And a detailed implementation is recommended for interested readers in the repository~\href{https://github.com/Zhu-Qianyu/PIV-FlowDiffuser}{https://github.com/Zhu-Qianyu/PIV-FlowDiffuser}.
\end{abstract}



\begin{keywords}
Particle image velocimetry  \sep Denoising Diffusion Model \sep Transfer-learning \sep Optical flow estimation 
\end{keywords}

\maketitle

\section{Introduction}
\label{sect_1}
Particle image velocimetry (PIV) provides instantaneous, quantitative, and whole-field velocity, and becomes a fundamental non-intrusive optical measurement technique in fluid mechanics ~\cite{adrian1984scattering,raffel2018particle,panciroli2015experiments,bardera2019wind,capone2021flow}. PIV works by introducing tracer particles into the flowing medium, illuminating them with a laser light sheet, and capturing two or more consecutive images of the particles with a camera. The instantaneous velocity vector field is estimated by analyzing the sequential particle recordings. 
However, the practical PIV images are often susceptible to degradation in the course of measurements due to factors such as nonuniform light illumination, light reflections, background noise sources, and camera dark noise~\cite{lee2022blind,lee2024surrogate}. Consequently, these effects impose stringent demands on both analytical accuracy and generalization~\cite{kahler2016main,sciacchitano2019uncertainty}.

Over the past 40 years, the cross-correlation and optical flow algorithms are the two primary algorithms employed for PIV processing~\cite{westerweel1997fundamentals,corpetti2006fluid,pan2015evaluating,zhong2017optical,wang2020globally,lee2021diffeomorphic,ai2025rethinking}. 
For example, the window deformation iterative method~(WIDIM) with central difference interrogation is verified to significantly improve the accuracy of estimation~\cite{wereley2001second,scarano2001iterativeimage}. 
An optimized surrogate image, replacing one raw image, can generate a more accurate and robust correlation signal when addressing image background interference~\cite{lee2024surrogate}. Despite their widespread use, cross-correlation methods extract velocity vectors from sparse interrogation windows, often resulting in low-resolution velocity fields. Optical flow methods estimate particle displacement based on the brightness preservation principle~\cite{corpetti2006fluid}, which assumes that the particle image brightness remains unchanged after movement. However, when the optical flow principle is violated---often due to image noise in practical measurements---the risk of failure significantly increases~\cite{scharnowski2020particle,liu2015comparison}.

To improve the velocity estimation, deep learning techniques have gained popularity in PIV applications in recent years. The basic idea is that neural networks are trained to predict vectors from two input images~\cite{lee2017piv,cai2019particle,zhang2020unsupervised,lagemann2021deep,yu2021effective,wang2022dense,cai2024physics,yu2024deep,reddy2025twins}. 
PIV-DCNN is the first regression deep convolutional neural network for PIV estimation. It employs a four-level structure and offers compromising results compared to the conventional cross-correlation methods~\cite{lee2017piv}. Since then, a series of deep learning models (PIV-NetS~\cite{cai2019dense}, PIV-LiteFlowNet~\cite{cai2019particle}, LightPIVNet~\cite{yu2021lightpivnet}, RAFT-PIV~\cite{lagemann2021deep}, PIV-PWCNet~\cite{zhang2023pyramidal}, etc.) have been proposed and evaluated with synthetic particle images.  Note that Cai et al. released a synthetic PIV dataset, which has been widely used by state-of-art methods for training deep learning models~\cite{cai2019dense}.
Similar to WIDIM, neural networks equipped with iterative updates often achieve better results. The iterative implementation can take various forms depending on the researchers, such as cascaded refinement~\cite{lee2017piv}, coarse-to-fine pyramids~\cite{zhang2023pyramidal}, and convolutional gated recurrent units (Conv-GRU)~\cite{lagemann2021deep}. 
Note that these models typically employ only a few iterations, with RAFT---the most iterative among them---usually set to 16 iterations.
Nevertheless, these deep learning-based PIV algorithms estimate dense velocity fields from particle images in an end-to-end manner, which improve the computational efficiency, accuracy, and spatial resolution compared to conventional cross-correlation or optical flow methods. 

\begin{figure}
\centering
\includegraphics[width=\columnwidth]{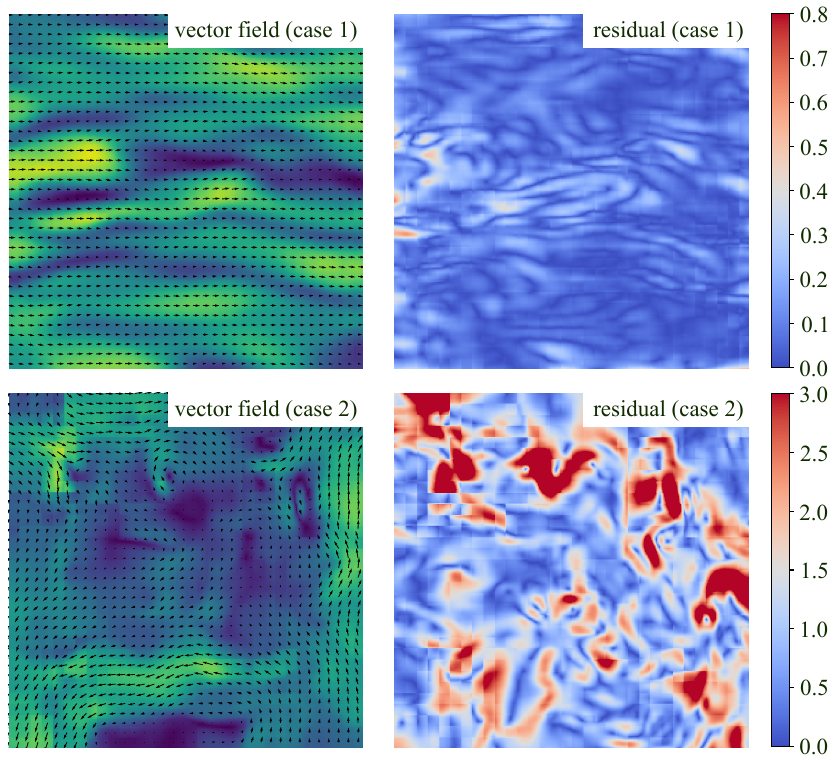}
	\caption{Results of the RAFT256-PIV method~\cite{lagemann2021deep} on two test cases. The left gives the vector fields, while the right part presents corresponding error maps. Note that some special error patterns are observed in the residuals, which could be further reduced with noise removal.}
	\label{fig_1}
\end{figure}

Despite the advancements of neural networks, their performance in practical measurements is not always satisfactory,
as illustrated in Fig.~\ref{fig_1}. Specifically, noticeable error patterns can be observed in the residual of estimations. The underlying reasons might be twofold. First, the architecture of existing PIV neural networks lacks an explicit correction mechanism, meaning that errors accumulated during iterative updates cannot be effectively mitigated. That says, the flow correction  $\Delta \boldsymbol{v}$ should be explicit modeled as a function w.r.t current estimation $\boldsymbol{v}$, i.e.,  $\Delta \boldsymbol{v}= \boldsymbol{f}_{update}(\boldsymbol{v}, \cdot)$. We argue that this explicit correction mechanism has the potential to rectify error patterns in the estimated flow field, thereby improving overall accuracy. Second, those neural models mentioned above are trained on synthetic datasets, which may differ significantly from the real-world test cases,  a discrepancy commonly referred to as the domain gap~\cite{tobin2017domain}.
The currently available open-access synthetic datasets often lack sufficient diversity, limiting deep models' ability of the generalization on practical applications~\cite{cai2019particle,zhang2023pyramidal}. Thus, in this work, we focus on the challenges of designing an explicit error correction network architecture and developing effective training strategies.

Denoising diffusion models are known for their ability to generate high-quality images through a sequence of iterative denoising operations, i.e.,  $p_\theta(\boldsymbol{x}_{t-1}|\boldsymbol{x}_t)$~\cite{dhariwal2021diffusion,ho2020denoising}. Our insight is that the denoising diffusion model meets the explicit correction requirement for more accurate PIV measurement. Meanwhile, a denoising diffusion model (FlowDiffuser~\cite{luo2024flowdiffuser}) has established a new benchmark in the task of optical flow estimation, which further motivates us to use denoising diffusion models for PIV analysis. Unfortunately, the FlowDiffuser trained from scratch does not yield satisfactory results for PIV applications. One potential cause is that the synthetic dataset is insufficient for this data-hungry iterative model. 
However, transfer learning could effectively address data scarcity and improve generalization ability~\cite{zhao2024comparison,hu2024industrial}. As a transfer learning method, fine-tuning pre-trained models has already become popular for complex models~\cite{han2021pre}. The substantial requirement for labeled data in training accurate and robust PIV models has led us to investigate transfer learning (pre-trained models) as a potential solution.

As a result, we introduce PIV-FlowDiffuser, a denoising diffusion model trained using transfer learning, for particle image velocimetry. Our PIV-FlowDiffuser is designed to reduce measurement errors through iterative denoising, with the expectation of improving the accuracy of velocity estimation. Additionally, we transfer the initial weights from a pre-trained optical flow model and fine-tune it on the PIV datasets. This fine-tuning approach, leveraging pre-trained parameters, enhances generalization and promotes efficient training. The main contributions are as follows:
\begin{enumerate}
\item The denoising diffusion model (PIV-FlowDiffuser) is applied to PIV for the first time. PIV-FlowDiffuser employs an explicit correction mechanism that refines the estimated flow field iteratively, thereby reducing overall measurement errors.

\item Transfer learning technique is employed to train deep models for PIV prediction, i.e., fine-tuning a pre-trained FlowDiffuser model from the computer vision domain. This approach leverages the robust feature extraction and flow reconstruction capabilities obtained through pre-training, potentially reducing the need for extensive labeled PIV data. Consequently, it enables accurate PIV estimation with reduced training time and enhances the model's generalization performance.

\item The feasibility of our PIV-FlowDiffuser model was validated through extensive synthetic and practical PIV images. Compared to the existing RAFT256-PIV model, our PIV-FlowDiffuser demonstrates decreased measurement errors, as evidenced by lower residuals in visualizations. Specifically, it achieves a $59.4\%$  reduction in average end-point error~(AEE). Furthermore, the model exhibits superior accuracy on previously unseen data, indicating excellent generalization performance.

\end{enumerate} 
The rest of this article is organized as follows. Section.~\ref{sect_2} details PIV-FlowDiffuser framework and transfer learning-based training. Section.~\ref{sect_3} outlines the experimental settings of evaluation. Section.~\ref{sect_4} presents the experimental results, including a thorough comparison with other baseline methods on both synthetic and practical PIV recordings. Section.~\ref{sect_5} gives a short conclusion of this work.

\section{Transfer-learning-based denoising diffusion models for PIV}
\label{sect_2}
In this section, we revisit the FlowDiffuser model~\cite{luo2024flowdiffuser} and present a straightforward adaptation method for PIV analysis. Additionally, we detail the transfer learning approaches employed to train a PIV model, including the selection of loss functions and training data.

\subsection{The FlowDiffuser model for PIV}

\begin{figure*}
\centering
\centering
\subfloat[]{\includegraphics[width=0.75\textwidth]{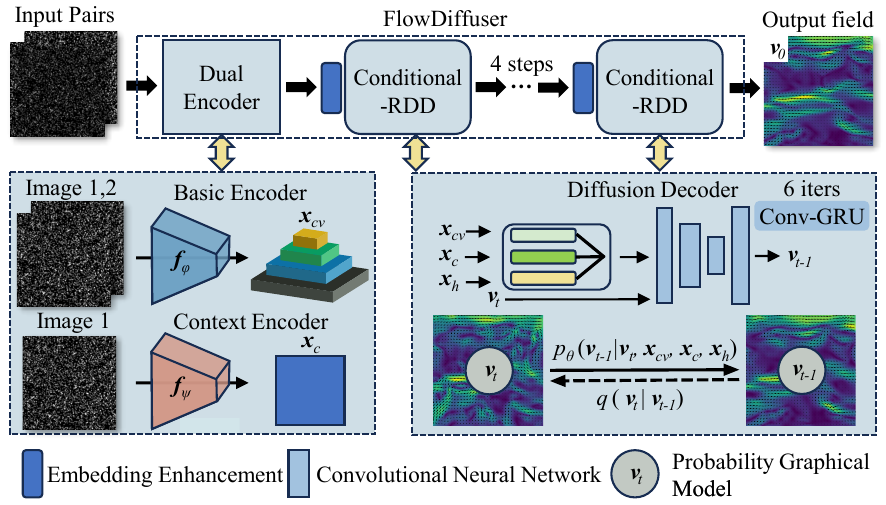}}
\hfill
\subfloat[]{\includegraphics[width=0.75\textwidth]{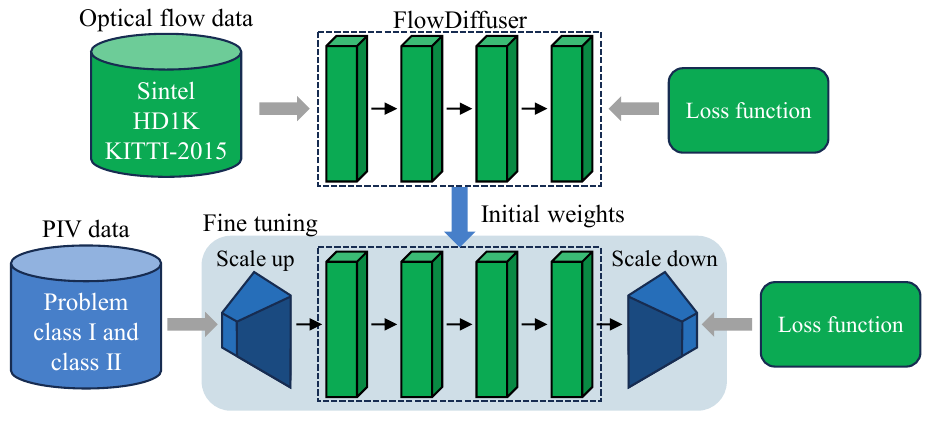}}
\caption{(a) The FlowDiffuser~\cite{luo2024flowdiffuser} includes two encoders (basic encoder and context encoder) and a series of conditional recurrent denoising decoder~(RDD). (b) The training method for PIV-FlowDiffuser. The initial weights are from the pre-trained model, and a simple adaptation module (scale up \& scale down) is adopted to better predict small-scale turbulence. The entire model was subsequently fine-tuned using a PIV-specified dataset.}
\label{fig_2}
\end{figure*}

Given the image pair $(I_1, I_2)$, the velocity estimation procedure can be formulated as $\boldsymbol{v}=f(\boldsymbol{I}_1, \boldsymbol{I}_2)$, where $f(\cdot)$ denotes the mapping from image pairs to velocity vectors~\cite{lee2017piv}. Iterative update is a common trick to effectively improve the performance of conventional PIV processing, i.e.,  $\boldsymbol{v}^{new} = \boldsymbol{v}^{old} +\Delta \boldsymbol{v}$, where $\Delta \boldsymbol{v}$ is the estimated corrector~\cite{lee2021diffeomorphic,lagemann2021deep,scarano2001iterativeimage}.
%
Our insight is that, this deterministic approach can be extended to estimate the conditional probability distribution $p(\boldsymbol{v}|\boldsymbol{I}_1, \boldsymbol{I}_2)$, providing a probabilistic generative perspective of the velocity field given the observed images. Similarly, the iterative update also has corresponding probabilistic formulation $p(\boldsymbol{v}^{new}|\boldsymbol{v}^{old},\boldsymbol{I}_1, \boldsymbol{I}_2)$. While different formulations remain fundamentally unchanged, adopting a probabilistic perspective enables the integration of conditional generative models into PIV measurements. 

As a powerful generative method, denoising diffusion models~\cite{ho2020denoising} are capable of synthesizing highly realistic images $\boldsymbol{x}_0$ via iterative backward diffusion (denoising), i.e., $p_\theta(\boldsymbol{x}_{t-1}|\boldsymbol{x}_{t})$.  The key idea is that a neural network ---parametrized by $\theta$---takes in two arguments $\boldsymbol{x}_t, t$, and outputs a vector $\mu_\theta(\boldsymbol{x}_t,t)$ and a matrix $\Sigma_\theta(\boldsymbol{x}_t,t)$. That says,  
\begin{equation}
    p_\theta(\boldsymbol{x}_{t-1}|\boldsymbol{x}_{t}) = \mathcal{N}(\boldsymbol{x}_{t-1}|\mu_\theta(\boldsymbol{x}_t,t),\Sigma_\theta(\boldsymbol{x}_t,t))
\end{equation}
where the initial $p_\theta(\boldsymbol{x}_T)=\mathcal{N}(\boldsymbol{0}, \boldsymbol{I})$ is diagonal Gaussian distribution. 
Note that the time step $t$ is crucial for encoding the denoising process and is explicitly incorporated into the model. Inspired by the proven effectiveness of denoising diffusion probabilistic models, FlowDiffuser~\cite{luo2024flowdiffuser} considers the conditional influence of the input images, i.e., $p_\theta(\boldsymbol{v}_{t-1}|\boldsymbol{v}_{t},\boldsymbol{I}_1, \boldsymbol{I}_2)$, to redefine optical flow estimation as a generative process. 
In addition, the FlowDiffuser model demonstrates strong alignment with the probabilistic formulation of PIV estimation.

To this end, we have established an overview of the FlowDiffuser concept. We now explain our interpretation and implementation of this approach, focusing on three key components: the Dual Encoder for enhanced image representation, embedding enhancement techniques for encoding the iterative step information, and the Conditional Recurrent Denoising Decoder (Conditional-RDD) for iterative updates, as illustrated in Fig.~\ref{fig_2}. 1) Dual encoders pre-process the input image pair ($\boldsymbol{I}_1$, $\boldsymbol{I}_2$), generating a context feature $\boldsymbol{x}_c$ and a $4D$ correlation volume~$\boldsymbol{x}_{cv}$, as given by the RAFT~\cite{teed2020raft}. The 4D correlation volumes can capture correlations of particle motion. Anyway, it means that the condition on particle images can be replaced with more representative features, i.e., $p_\theta(\boldsymbol{v}_{t-1}|\boldsymbol{v}_{t},\boldsymbol{x}_c, \boldsymbol{x}_{cv})$. 2) The embedding enhancement module~\cite{luo2024flowdiffuser} explicitly synchronizes the time steps $t$ to the flow decoder. That says, $\boldsymbol{x}_o = EE(t,\boldsymbol{v}_t,\boldsymbol{x}_c,\boldsymbol{x}_{cv})$ enhances the temporal embedding features, and is fed into the GRU and flow head modules.
3) The conditional recurrent denoising decoder (conditional-RDD) takes the initial noisy flow field $\boldsymbol{v}_t$ as input and optimizes it to $\boldsymbol{v}_{t-1}$ through a denoising step (recurrent Conv-GRU neural network). Besides the flow field $\boldsymbol{v}_t$ is explicitly conditioned, the hidden feature $\boldsymbol{x}_h$  is also considered as an alternative representation of the flow field. Hence, we define the implemented denoising decoder as $p_\theta(\boldsymbol{v}_{t-1}|\boldsymbol{v}_{t},\boldsymbol{x}_c, \boldsymbol{x}_{cv}, \boldsymbol{x}_h)$,  where $p_\theta$ is the symbolic representation of the model and $\theta$ is the learnable weights. During inference, noise is progressively removed from the initial noisy estimation $\boldsymbol{v}_T$, which follows a standard Gaussian distribution. These three components collectively contribute to the robustness and accuracy of the FlowDiffuser model in optical flow tasks~\cite{luo2024flowdiffuser}.

To further enhance the model's ability to measure small-scale turbulence, we employ a straightforward trick: upsampling the input images by a factor of 2 using bilinear interpolation (Fig.~\ref{fig_2}). By increasing the image resolution through upsampling, the neural network can better capture fine details and intricate patterns associated with small-scale turbulence. Note that a corresponding downsampling process is performed at the output stage to maintain the same shape for the output flow field. This adaptation helps improve the measurement accuracy of multiscale flow fields.

\subsection{Transfer-learning based training}
In this part, we present the training method for FlowDiffuser model by employing a transfer learning (fine-tuning)~\cite{oquab2014learning,chang2017unsupervised}. It leverages the robust features and latent flow representation learned during pre-training on large-scale optical flow datasets, which provides a good initialization for our PIV analysis. By fine-tuning the pre-trained model, we adapt it to estimate both global and local flow characteristics with minimal domain-specific data. This strategy significantly reduces training time and computational resources. 

To adapt the FlowDiffuser model for PIV applications, structural adaptations are used to bridge the source domain (color sequence) and our target PIV data, as shown in Fig.~\ref{fig_2}. First, the input PIV images, originally in grayscale, are transformed into a pseudo-color format by replicating the single channel across three channels. This conversion allows the model to leverage pre-trained features from color image datasets. Additionally, the output flow fields are downsampled to match the desired resolution, along with the upsampling trick for images. With these adaptations, the PIV application of transfer learning to the FlowDiffuser model becomes feasible.

During the fine-tuning process, all the model parameters are updated (none are frozen) since the input images and output velocity fields belong to different domains in the transfer learning framework. We also utilize advanced loss functions to guide the training process, given as,
\begin{equation}
\mathcal{L}=\mathop{\mathbb{E}}_{\boldsymbol{v}_0\sim q(\boldsymbol{v}_0|c),t\sim[1,T]}\|\boldsymbol{v}_{gt}-{\boldsymbol{v}}_0\|_1
\label{eq:6}
\end{equation}
where $\|\cdot\|_1$ indicates the $l_1$ distance between flow ground truth $\boldsymbol{v}_{gt}$ and conditional denoising result ${\boldsymbol{v}}_0$, and $c$ is the abbreviation of the aforementioned probabilistic conditions.
A one-cycle learning rate scheduler~\cite{smith2019super} is used to refine the network, facilitating precise adaptation to the PIV domain.

\subsection{Datasets}
\label{subsect_dataset}
The transfer learning framework involves a source domain dataset for initial model training and a target domain dataset for fine-tuning the pre-trained model. Additionally, test datasets are also required for performance evaluation. Here, we briefly introduce these three datasets used in this work. Regarding the source domain dataset, FlowDiffuser models~\cite{luo2024flowdiffuser} are pre-trained on a combination of Sintel, KITTI-2015~\cite{menze2015object}, and HD1K. 
Specifically, the officially released FlowDiffuser model\footnote{https://github.com/LA30/FlowDiffuser} is employed as the pre-trained model for PIV analysis. 

Regarding the target domain, two synthetic datasets are considered: Problem Class 1\footnote{https://github.com/shengzesnail/PIV\_dataset} and Problem Class 2\footnote{https://codeocean.com/capsule/7226151/tree/v1}~\cite{lagemann2021deep}. Problem Class 1, introduced by Cai et al.~\cite{cai2019dense}, consists of synthetic particle image pairs with corresponding ground-truth fluid motions, designed to resemble ideal experimental conditions with high particle density and intensity. The Problem Class 1 dataset contains $14,150$ particle image pairs at a resolution of $256 \times 256 $. It includes six fluid cases~(uniform channel flow, direct numerical simulation of isotropic turbulence, flows around a backward-facing step,  two-dimensional flows past a cylinder, DNS of turbulent channel flow, and simulations of a sea surface flow driven by an SQG model). This dataset contains over $10$ different operating conditions, with an average of about $1,000$ image samples per condition. The Problem Class 1 dataset is divided into training, validation, and test sets in an $8:1:1$ ratio. Problem Class 2 is based on the same ground-truth flow fields as Problem Class 1 but with reduced particle density, intensity, and overall signal-to-noise ratio to better model real experimental data conditions. The Problem Class 2 dataset consists of $19,000$ training image pairs and $1,000$ validation image pairs.  Among these $1,000$ cases, we classify $224$ as Backstep, $300$ as JHTDB, $92$ as DNS-Turbulence, $128$ as Cylinder, $99$ as SQG, $57$ as Uniform, and the remaining $100$ unrecognized cases as Other.

To further evaluate the performance in complex flow fields, we test the PIV models on a turbulent wavy channel flow (TWCF) case~\cite{rubbert2019streamline,lagemann2021deep}. This test case consists of experimental PIV measurements in an Eiffel-type wind tunnel and features a high spatial resolution of $2160 \times 2560$ pixels.  
Moreover, due to the waviness of the sidewall (sinusoidal shape), both locally adverse and pressure gradient flow conditions occur in the channel, making this case well-suited for evaluating model performance on real PIV images.

\section{Experimental arrangement}
\label{sect_3}
In this section, we outline the experimental setup to evaluate performance through a series of experiments on both synthetic and real PIV cases. Specifically, three scenarios are considered: (1) accuracy assessment on synthetic PIV datasets where the training and testing data share the same distribution, (2) generalization assessment on synthetic PIV datasets where the training and testing data belong to different domains, and (3) practical performance on experimental PIV recordings with unknown distributions. Besides, the training and inference costs of multiple deep-learning models are also tested. Before presenting the results of these experiments,  the performance metrics for PIV models are introduced, along with the baseline methods used for comparison and other experimental details.

\subsection{Evaluation metrics}
To visually assess the performance, we provide the error residual map, calculated as $\|\boldsymbol{v}_{0}-\boldsymbol{v}_{gt}\|_2$, for intuitive evaluation. Besides, given the availability of ground truth data, we also employ several classic quantitative metrics to evaluate performance: average end-point error~(AEE)~\cite{lagemann2021deep}, root mean square error~(RMSE)~\cite{raffel2018particle,lee2017piv},  and average angular error~(AAE)~\cite{sharmin2012optimal,lu2021accurate}.
\begin{equation}
\label{eq:7}
\begin{split}
&AEE=\frac{1}{N}\sum_{i=1}^{N}\|\boldsymbol{v}_{0}^{(i)}-\boldsymbol{v}_{gt}^{(i)}\|_2\\
&RMSE = \sqrt{\frac{1}{N} \sum_{i=1}^N \| \boldsymbol{v}_{pred}^{(i)} - \boldsymbol{v}_{gt}^{(i)} \|_2^2}\\
&AAE=\frac{1}{N}\sum_{i=1}^{N}\arccos\left(\frac{\boldsymbol{v}_{0}^{(i)}\cdot \boldsymbol{v}_{gt}^{(i)}}{\|\boldsymbol{v}_{0}^{(i)}\|_2\cdot\|\boldsymbol{v}_{gt}^{(i)}\|_2}\right)
\end{split}
\end{equation}
where $\boldsymbol{v}_{gt}^{(i)}$ and $\boldsymbol{v}_{0}^{(i)}$ denotes the $i^{th}$ ground truth and the estimated velocity vector respectively. $N$ is the total number of pixels and the $\|\cdot\|_2$ refers to the Euclidean distance.

\subsection{Baseline methods}

The baseline methods include several classical PIV analysis approaches based on cross-correlation and deep neural networks, such as the WIDIM~\cite{scarano2001iterativeimage}, PIV-DCNN~\cite{lee2017piv}, PIV-LiteFlowNet~\cite{hui2018liteflownet}, RAFT256-PIV~\cite{lagemann2021deep}, RAFT32-PIV~\cite{lagemann2021deep}, FlowFormer-PIV~\cite{huang2022flowformer} and Twins-PIVNet~\cite{reddy2025twins}.

Since two datasets (Problem Class 1 and Problem Class 2, detailed in Section.~\ref{subsect_dataset}) are used for training different FlowDiffuser models, we incorporate the dataset name into our model's abbreviation. As a result, we define two primary PIV-FlowDiffuser models: PIV-FlowDiffuser-class1 and PIV-FlowDiffuser-class2. This suffix-based naming rule is also applied to other algorithms~(RAFT256-PIV and FlowFormer-PIV), resulting in RAFT 256-PIV-class1, RAFT 256-PIV-class2, FlowFormer-PIV-class1 and FlowFormer-PIV-class2.

Additionally, to conduct ablation studies, we introduce two variants: (1) To evaluate the contribution of our adaptation module, we remove the upsampling/downsampling adaptation module, resulting in a model denoted as PIV-FlowDiffuser-class1(*); (2) To assess the effect of transfer learning, we also employ a pre-trained model (FlowDiffuser) without any scale adaptation nor fine-tuning.

\section{Results and discussions}
\label{sect_4}
\subsection{Accuracy performance on synthetic images}
\label{sect_4_1}

\begin{table*}[width=\textwidth,cols=4,pos=!htb]
\caption{The AEE results evaluated on the Problem Class 1 dataset. The best results are highlighted in bold and the second is underlined. The data of WIDIM, PIV-DCNN, PIV-LiteFlowNet-en, RAFT256-PIV, and Twins-PIV are sourced from~\cite{lagemann2021deep,reddy2025twins}.  (unit: pixels per frame)}
\label{tbl1}
\begin{tabular*}{\tblwidth}{@{}L|CCCCC}
\hline
 Method & \makecell{Backstep} & \makecell{JHTDB-\\channel} & \makecell{DNS-\\turbulence} & Cylinder & SQG   \\
\hline
WIDIM & 0.034 & 0.084 & 0.304 & 0.083 & 0.457  \\
PIV-DCNN & 0.049 & 0.117 & 0.334 & 0.100 & 0.479  \\
PIV-LiteFlowNet-en & 0.033 & \underline{0.075} & 0.122 & 0.049 & 0.126  \\
RAFT256-PIV-class1 & 0.016 & 0.137 & 0.093 & \underline{0.014} & 0.117  \\
Twins-PIV-class1 & \underline{0.013} & 0.092 & \underline{0.056} & \textbf{0.012} & \underline{0.091} \\
PIV-FlowDiffuser-class1(*) & 0.041 & 0.111 & 0.148 & 0.091 & 0.208 \\
PIV-FlowDiffuser-class1 & \textbf{0.007} & \textbf{0.029} & \textbf{0.039} & 0.019 & \textbf{0.052} \\
\hline
\end{tabular*}
\end{table*}

\begin{table*}[width=\textwidth,cols=4,pos=!htb]
\caption{The AEE results evaluated on the Problem Class 2 dataset. The best results are highlighted in bold and the second is underlined. (unit: pixels per frame)}
\label{tbl2}
\begin{tabular*}{\tblwidth}{@{} L|CCCCCCC@{} }
\hline
Method & Backstep & \makecell{JHTDB-\\channel} & \makecell{DNS-\\Turbulence} & Cylinder & SQG & Uniform & Other  \\
\hline
RAFT256-PIV-class2 & \textbf{0.131} & \underline{0.476} & \underline{0.646} & \textbf{0.124} & \underline{0.593} & \textbf{0.174}  & \textbf{0.380} \\
FlowFormer-PIV-class2 & 0.377 & 0.680 & 1.316 & { 0.297} & 1.042 & 0.850  & 1.406  \\
PIV-FlowDiffuser-class2 & \underline{0.155} & \textbf{0.296} & \textbf{0.565} & \underline{0.138} & \textbf{0.466} & \underline{0.328} & \underline{0.587}  \\
\hline
\end{tabular*}
\end{table*}

\begin{figure*}[!htb]
\centering
\includegraphics[width=.92\textwidth]{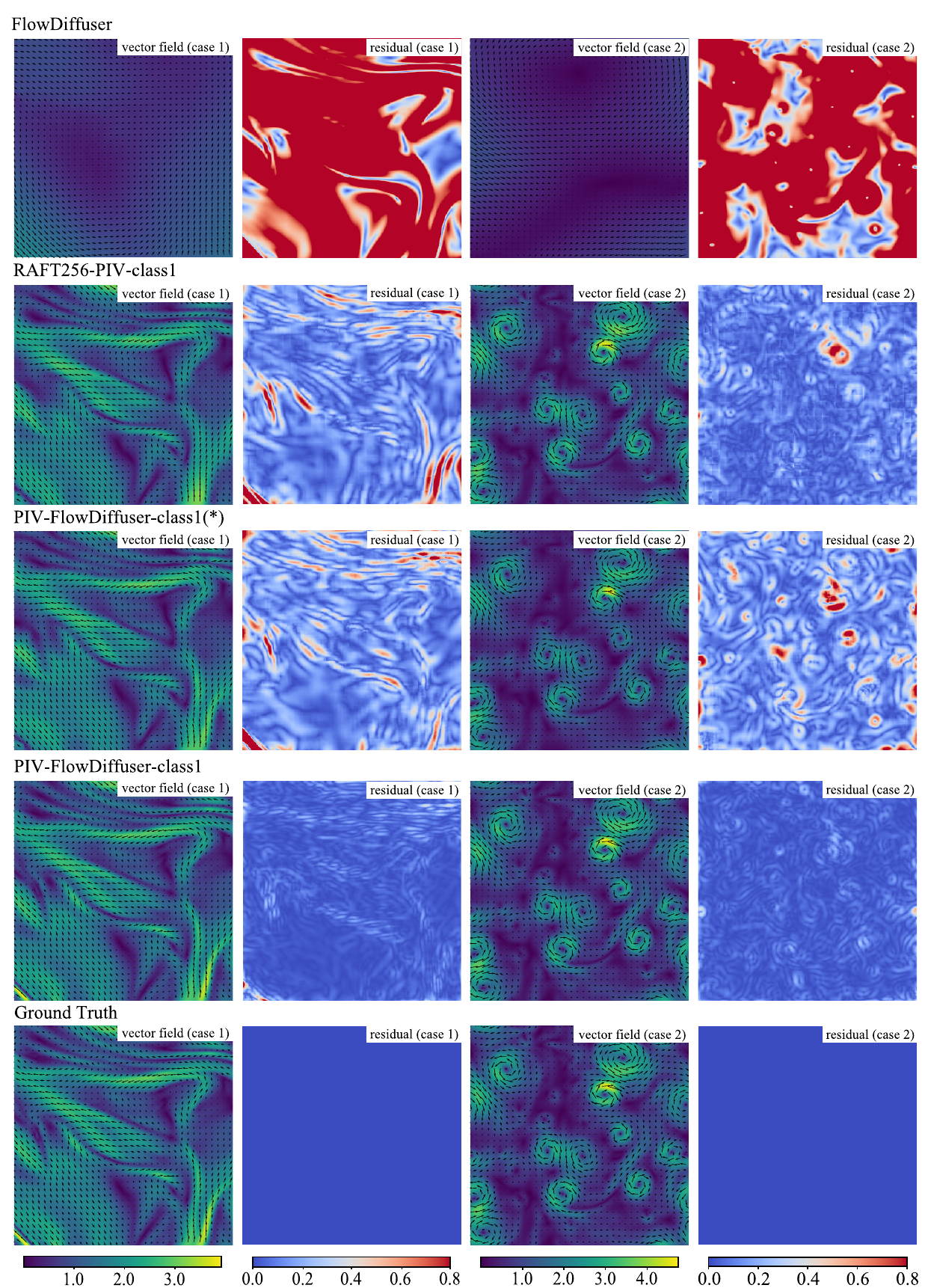}
\caption{Velocity fields and corresponding absolute residuals of Problem Class 1 computed by different methods. Two cases (left: JHTDB, right: SQG) are considered. The color backgrounds denote the corresponding velocity/residual magnitude. Best viewed in color. (unit: pixels per frame)}
\label{fig_3}
\end{figure*}

Fig.~\ref{fig_3}  presents the vector fields and corresponding error distributions of two cases from the test set of the Problem Class 1. 
The RAFT256-PIV-class1 method successfully captures the primary flow structures. However, it exhibits measurement errors, particularly in regions of small-scale turbulent flow.
The baseline pre-trained model (FlowDiffuser) fails to predict the flow accurately, whereas fine-tuning (PIV-FlowDiffuser-class1(*)) significantly reduces the measurement error, achieving performance comparable to RAFT256-PIV-class1. Furthermore, by integrating an adaptation module (upsampling and downsampling), PIV-FlowDiffuser-class1 surpasses both RAFT256-PIV-class1 and PIV-FlowDiffuser-class1(*). It means that the upsampling trick helps to resolve the small-scale turbulence structure, as expected.

Table.~\ref{tbl1} presents similar AEE results on the test set of Problem Class 1. Among the five test subsets, PIV-FlowDiffuser-class1 achieved the lowest measurement error in four subsets, consistent with the findings presented in Fig.~\ref{fig_3}. Note that PIV-FlowDiffuser-class1(*) algorithm also demonstrates acceptable performance compared to the traditional WIDIM baseline. The comparison between PIV-FlowDiffuser-class1 and PIV-FlowDiffuser-class1(*) reveals that the simple upsampling trick effectively enhances the accuracy of PIV measurements. Regarding the baseline methods, the recent Twins-PIV~\cite{reddy2025twins} method demonstrates the most competitive overall performance, achieving one top-ranking and three second-best results. As summarized in the Table.~\ref{tbl3}, our PIV-FlowDiffuser-class1 achieves a $59.4\%$ error reduction $(0.0352)$ compared to  the average error of RAFT256-PIV-class1 $(0.0866)$,  demonstrating improved measurement accuracy.

Table~\ref{tbl2} presents the AEE results on the test set of Problem Class 2. Since Problem Class 2 corresponds to a large displacement dataset, all methods exhibit larger errors compared to the data in Table~\ref{tbl1}. It implies that measurements with large particle displacements are associated with increased measurement errors. The errors might originate from the linear motion approximation of actual curved particle trajectories with large displacement --- a systematic error beyond the current research scope~\cite{scharnowski2013effect,lee2021diffeomorphic}. Without surprise, our PIV-FlowDiffuser-class2 reduces the RAFT error by half, achieving state-of-the-art performance.

\subsection{Generalization to out-of-domain dataset}
\begin{figure*}[!htb]
\centering
\includegraphics[width=.92\textwidth]{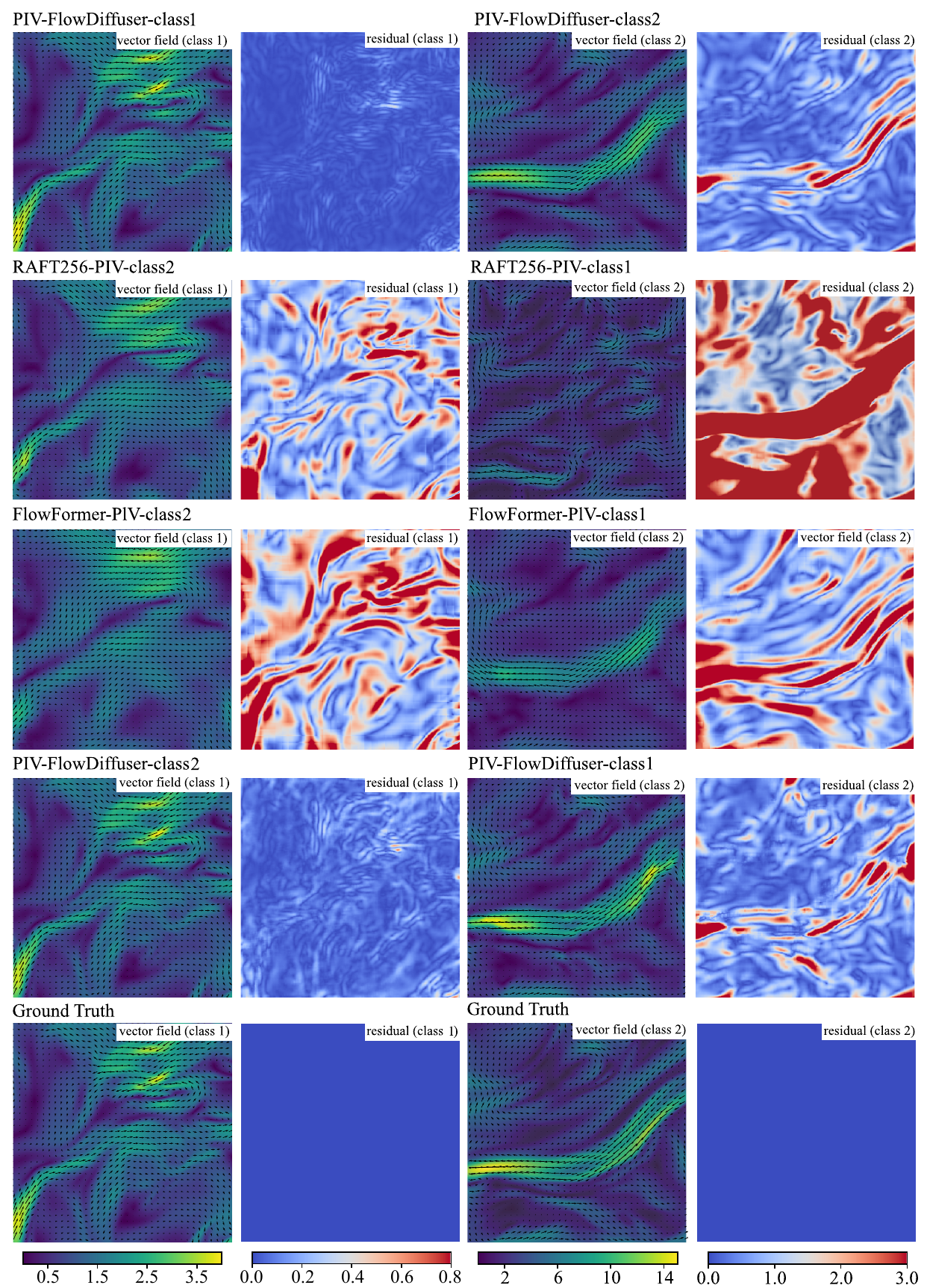}
\caption{Velocity fields and corresponding absolute residuals computed by different methods. Two cases (left: a JHTDB from Problem Class 1, right: a JHTDB from Problem Class 2) are considered. The color backgrounds denote the corresponding velocity/residual magnitude. Best viewed in color. (unit: pixels per frame)
}
\label{fig_4}
\end{figure*}

\begin{table*}[width=\textwidth,cols=4,pos=htb]
\caption{The AEE, RMSE and AAE results evaluated on out-of-domain dataset. The best results are highlighted in bold. Note that the in-domain results are also provided (with the best results shown in parentheses). (unit: pixels per frame)}
\label{tbl3}
\begin{tabular*}{\tblwidth}{@{} L|CCC|CCC @{} }
\hline
Method &  & Problem Class 1 & & &Problem Class 2&  \\
 & AEE & RMSE & AAE &  AEE &RMSE & AAE \\
\hline
RAFT256-PIV-class1 & 0.0866 & 0.0741 & 0.1631 & 4.7564 & 5.4709 & 1.6131 \\
FlowFormer-PIV-class1 & 0.3608 & 0.2956 & 0.3412 & 1.2518 & 1.4561 & 0.4236 \\
PIV-FlowDiffuser-class1 & (0.0352) & (0.0273) & (0.1283) & \textbf{0.5537} & \textbf{0.7279} & \textbf{0.2683} \\
\hline
RAFT256-PIV-class2 & 0.2107 & 0.1726 & 0.2644 & 0.3540 & 0.4502 & 0.2380 \\
FlowFormer-PIV-class2 & 0.4205 & 0.3483 & 0.3890 & 0.7662 & 0.9180 & 0.3365 \\
PIV-FlowDiffuser-class2 & \textbf{0.1021} &\textbf{ 0.0848} & \textbf{0.1667} & (0.3124) & (0.3921) & (0.1795) \\
\hline
\end{tabular*}
\end{table*}

In addition to the in-domain testing~(Sect.~\ref{sect_4_1}), out-of-domain generalization testing~\cite{wald2021calibration} is crucial for assessing robustness and reliability, and it might represent a promising avenue for developing widely deployable PIV models across diverse experimental conditions. By evaluating the deep neural models on datasets that differ significantly from the training data, researchers can identify potential weaknesses in handling diverse flow conditions, varying illumination, and different imaging noise levels.

Fig.~\ref{fig_4} illustrates the predicted vector fields (JHTDB flow) in scenarios where the test data exhibit significant distributional divergence from the training images. When evaluated on problem class 2 samples, the RAFT256-PIV-class1 model trained exclusively on class 1 data demonstrates unsatisfactory performance. Trained from scratch, RAFT256-PIV-class1 fails to produce accurate predictions for out-of-domain datasets, clearly manifesting the domain gap challenge as evidenced by the prediction errors. 
By contrast, the transfer learning-enhanced FlowFormer-PIV-class1 model achieves improved accuracy when handling Problem Class 2 cases, demonstrating that fine-tuning a pre-trained model substantially enhances cross-domain generalization capabilities. More significantly, our proposed PIV-FlowDiffuser framework outperforms all counterparts in cross-domain scenarios, exhibiting exceptional robustness (as evidenced by quantitative metrics in Table.~\ref{tbl3}) and statistically reliable predictions across all test conditions.

As expected, Table.~\ref{tbl3} statistically reveals significant performance disparities between in-domain and out-of-domain configurations. Note that our PIV-FlowDiffuser framework maintains acceptable performance despite being exclusively trained on single-domain datasets. These experimental statistics confirm the enhanced generalization capacity obtained by transfer learning (fine-tuning).

\subsection{Performance on practical PIV images}
\begin{figure*}[!htb]
	\centering
	\includegraphics[width=.92\textwidth]{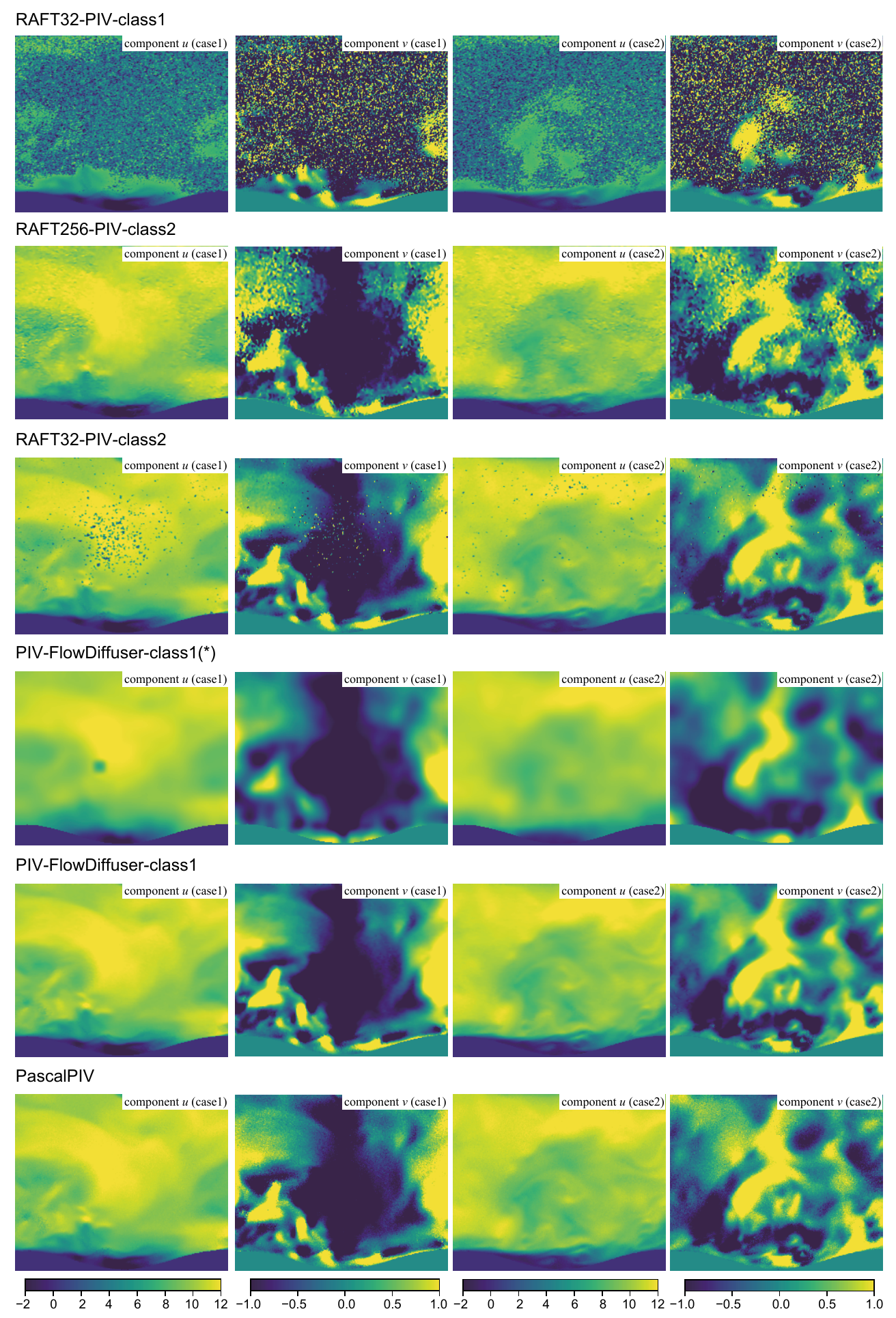}
	\caption{Two cases of TWCF data are visualized with separate velocity components. The color backgrounds denote the corresponding component value. Best viewed in color. (unit: pixels per frame)
    }
	\label{fig:5}
\end{figure*}

Fig.~\ref{fig:5} presents different results for two cases from the practical turbulent wavy channel flow (TWCF) dataset. Given the absence of ground velocity truth, the PascalPIV method implements a multigrid cross-correlation algorithm to produce satisfactory velocity field estimates, thus making its outputs suitable as the current gold standard~\cite{lagemann2021deep}. The flow fields exhibit multiscale fluctuations that span an order of magnitude in amplitude variation, providing benchmark cases for evaluating different estimation algorithms on practical PIV images.  
RAFT32-PIV-class1 fails in two TWCF cases (with maximum velocities of 12 pixels/frame), primarily attributable to the substantial domain shift between the training dataset and testing cases. That says, the maximum displacement of the training set (Problem Class 1) is less than $10$ pixels/frame. To mitigate this problem, domain expansion (data augmentation) is a straightforward strategy that addresses the observed inter-domain discrepancy. Trained on an expanded dataset (Problem Class 2), RAFT256-PIV-class2 and RAFT32-PIV-class2 generate acceptable estimations, despite the presence of some outliers.

As an alternative solution, transfer learning can use relevant optical flow knowledge learnt from natural images to generalizable PIV analysis. With an initialized pre-trained model,  PIV-FlowDiffuser-class1(*) also produces reasonable measurements, demonstrating the generalization ability of transfer learning. Meanwhile, its results exhibit some non-negligible errors in the area with small, complex turbulent structures. By further incorporating the adaptation module (upsampling trick), PIV-FlowDiffuser-class1 reduces the prediction noise significantly,  resulting in comparable performance to PascalPIV and outperforming other PIV analysis baselines.

\subsection{Experiments on computational cost }

\begin{table*}[width=\textwidth,cols=4,pos=h]
\caption{The training time and average inference time for different methods. The data of RAFT256-PIV is sourced from~\cite{lagemann2021deep}, and Twins-PIV sourced from~\cite{reddy2025twins}. }
\label{tbl4}
\begin{tabular*}{\tblwidth}{@{} L|CCC @{} }
\hline
Method & Device & Training-time(h)  & Average inference time(sec) \\
\hline
WIDIM                   & -                          & -     & 0.86 \\
PIV-DCNN                & -                          & -     & 2.23 \\
PIV-LiteFlowNet-en      & -                          & -     & 0.13 \\
RAFT256-PIV             & Two NVIDIA Quadro RTX 6000 GPUs    & $\sim$18 (from scratch)    & 0.08 \\
Twins-PIV-class1        & Two NVIDIA Quadro RTX 8000 GPUs    & $\sim$21 (from scratch)    & 0.08 \\
Twins-PIV-class2        & Two NVIDIA Quadro RTX 8000 GPUs    & $\sim$32  (from scratch)   & 0.08 \\
FlowFormer-PIV-class1   & One NVIDIA GeForce RTX 3090 GPU   & $\sim$0.5 (fine-tuning)   & 0.05 \\
FlowFormer-PIV-class2   & One NVIDIA GeForce RTX 3090 GPU  & $\sim$1.5 (fine-tuning)   & 0.05 \\
PIV-FlowDiffuser-class1 & One NVIDIA GeForce RTX 3090 GPU  & $\sim$2.0 (fine-tuning)   & 0.27 \\
PIV-FlowDiffuser-class2 & One NVIDIA GeForce RTX 3090 GPU  & $\sim$5.0 (fine-tuning)   & 0.27 \\
\hline 
\end{tabular*}
\end{table*}

Table 4 presents the training time and inference time for different neural networks of the PIV application.  Compared to the recent Twins-PIV models trained from scratch, the training time (fine-tuning) of PIV-FlowDiffuser is reduced by $84.4\%\sim90.5\%$. 
It verified that transfer learning can significantly save training time in building an outstanding PIV estimation neural network.
Due to the increased complexity of the dataset (Problem Class 2), PIV-FlowDiffuser-class2 exhibits a slower convergence rate and requires more training time ($\sim5h$). Nevertheless, this is still more efficient than training from scratch.
In terms of inference time, PIV-FlowDiffuser is much slower than Twins-PIV and RAFT256, because it integrates an adaptation module, which upscales the image by a factor of two. 
Considering the accuracy improvement, we argue that the cost of PIV-FlowDiffuser is affordable given the current computing hardware.

\section{Conclusion}
\label{sect_5}
In this work, we propose PIV-FlowDiffuser, a novel framework that introduces denoising diffusion models into PIV analysis. By iteratively performing conditional denoising, PIV-FlowDiffuser progressively refines the predicted flow fields, effectively suppressing noise and capturing small-scale turbulent structures. This leads to accurate velocity estimation and fewer outliers, outperforming conventional deep learning-based baselines.
Secondly, we use transfer learning through a fine-tuning strategy, adapting a pre-trained optical flow model to the PIV domain. This significantly reduces the training time while enhancing the model's generalization ability, even under domain shifts. The fine-tuned model benefits from generic motion features learned from large-scale datasets, enabling efficient adaptation with limited PIV-specific data.
Besides, extensive experiments on both synthetic datasets and real-world PIV images confirm the outstanding accuracy and robustness of our PIV-FlowDiffuser. Notably, the method demonstrates strong generalization to out-of-domain scenarios, highlighting its practical potential for diverse PIV applications beyond the training distribution.
Moving forward, we believe that the integration of denoising diffusion models into PIV will pave the way for new directions in accurate flow measurement, particularly under challenging turbulent and unsteady flow conditions.

\bibliographystyle{model1-num-names}
\bibliography{main}

\end{document}